\documentclass[sigconf]{acmart}

\usepackage{booktabs} 
\usepackage{amssymb} 
\usepackage{graphicx} 
\usepackage{amsmath}
\usepackage{hyperref}
\usepackage{textcomp}
\usepackage{multirow}
\usepackage{mathtools}
\usepackage{dsfont}

\setcopyright{rightsretained}

\acmDOI{10.475/123_4}

\acmISBN{123-4567-24-567/08/06}

\acmConference[WOODSTOCK'97]{ACM Multimedia conference}{July 1997}{El Paso, Texas USA}
\acmYear{1997}
\copyrightyear{2016}

\acmArticle{4}
\acmPrice{15.00}

\begin{document}
\title{Deep Differential Recurrent Neural Networks}

\author{Naifan Zhuang}
\affiliation{%
  \institution{University of Central Florida}
  \streetaddress{P.O. Box 1212}
  \city{Orlando}
  \state{FL}
  \postcode{43017-6221}
}
\email{zhuangnaifan@knights.ucf.edu}

\author{The Duc Kieu}
\affiliation{%
  \institution{University of the West Indies}
  \streetaddress{P.O. Box 1212}
  \city{St. Augustine}
  \state{Trinidad and Tobago}
  \postcode{43017-6221}
}
\email{ktduc0323@yahoo.com.au}

\author{Guo-Jun Qi}
\affiliation{%
  \institution{University of Central Florida}
  \streetaddress{1 Th{\o}rv{\"a}ld Circle}
  \city{Orlando}
  \country{FL}}
\email{guojunq@gmail.com}

\author{Kien A. Hua}
\affiliation{%
  \institution{University of Central Florida}
  \streetaddress{1 Th{\o}rv{\"a}ld Circle}
  \city{Orlando}
  \country{FL}}
\email{kienhua@cs.ucf.edu}







\renewcommand{\shortauthors}{Zhuang et al.}

\begin{abstract}


Due to the special gating schemes of Long Short-Term Memory (LSTM), LSTMs have shown greater potential to process complex sequential information than the traditional Recurrent Neural Network (RNN). 
The conventional LSTM, however, fails to take into consideration the impact of salient spatio-temporal dynamics present in the sequential input data.
This problem was first addressed by the differential Recurrent Neural Network (dRNN), which uses a differential gating scheme known as Derivative of States (DoS).
DoS uses higher orders of internal state derivatives to analyze the change in information gain caused by the salient motions between the successive frames. The weighted combination of several orders of DoS is then used to modulate the gates in dRNN. 
While each individual order of DoS is good at modeling a certain level of salient spatio-temporal sequences, the sum of all the orders of DoS could distort the detected motion patterns. 
To address this problem, we propose to control the LSTM gates via individual orders of DoS and stack multiple levels of LSTM cells in an increasing order of state derivatives. 
The proposed model progressively builds up the ability of the LSTM gates to detect salient dynamical patterns in deeper stacked layers modeling higher orders of DoS,
and thus the proposed LSTM model is termed deep differential Recurrent Neural Network ($d^2$RNN).
The effectiveness of the proposed model is demonstrated on two publicly available human activity datasets: NUS-HGA and Violent-Flows. 
The proposed model outperforms both LSTM and non-LSTM based state-of-the-art algorithms. 

\end{abstract}



\maketitle

\section{Introduction}


Recent years have witnessed a revival of the Long Short-Term Memory (LSTM) \cite{hochreiter1997long}, thanks to the special gating mechanism that controls access to memory cells. The superior capability of LSTM has been shown in a wide range of problems such as machine translation \cite{sutskever2014sequence, bahdanau2014neural}, speech recognition \cite{graves2013speech}, and multi-modal translation \cite{venugopalan2014translating}. Compared with many existing spatio-temporal features \cite{klaser2008spatio, scovanner20073} from the time-series data, LSTM uses either a hidden layer \cite{schuster1997bidirectional} or a memory cell \cite{hochreiter1997long} to learn the time-evolving states which model the underlying dynamics of the input sequences. 
In contrast to the conventional RNN, the major component of LSTM is the memory cell which is modulated by three gates: input, output, and forget gates. These gates determine the amount of dynamic information entering/leaving the memory cell. The memory cell has a set of internal states, which store the information obtained over time. In this context, these internal states constitute a representation of an input sequence learned over time.

LSTMs have shown tremendous potential in activity recognition tasks \cite{donahue2015long, baccouche2010action, grushin2013robust}. The existing LSTM model represents a video by integrating all the available information from each frame over time. It was pointed out in \cite{veeriah2015differential} that for an activity recognition task, not all frames contain salient spatio-temporal information which is equally discriminative to different classes of activities. Many frames contain non-salient motions which are irrelevant to the performed actions. Since the gate units in LSTM do not explicitly consider whether a frame contains salient motion information when they modulate the input and output of the memory cells, LSTM is insensitive to the dynamical evolution of the hidden states given the input video sequences and cannot capture the salient dynamic patterns. dRNN addresses this problem and models the dynamics of actions by computing different orders of Derivative of State (DoS). DoS models the change in information gain caused by the salient motions between the successive frames using higher orders of internal state derivatives. Intuitively, 
1st-order DoS represents the velocity of change of internal state memory while 2nd-order DoS represents the acceleration of memory state change.
This reveals that the conventional LSTM, whose internal cell is simply 0th-order DoS, only captures the locality of information change.

\begin{figure*}
	\centering
		\includegraphics[width=1.0\linewidth]{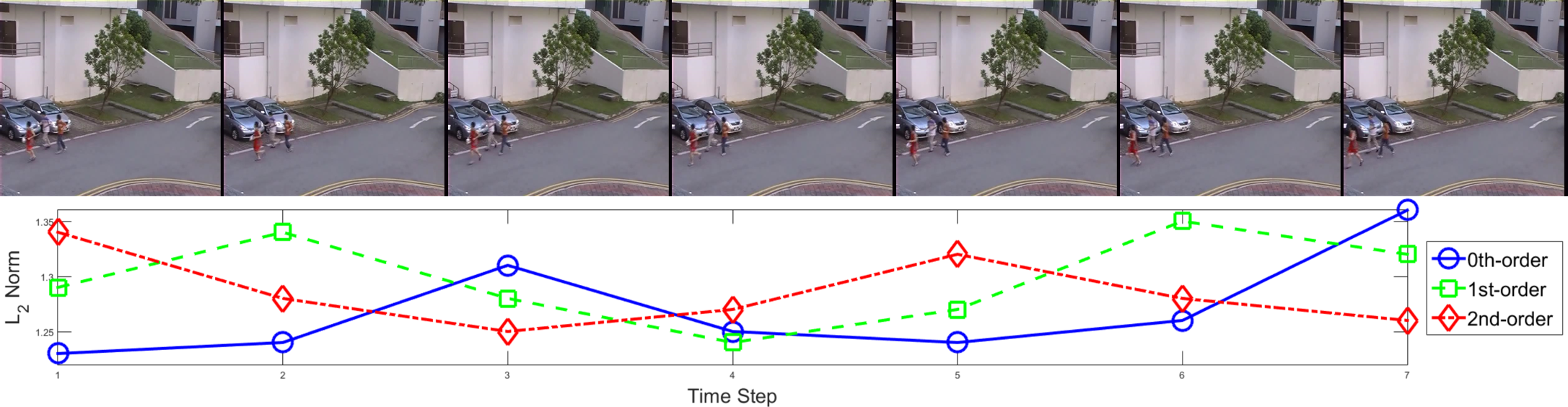}
	\caption{The energy curves of the 0th-, 1st-, and 2nd-orders of DoS over an example of sequence for the activity "RunInGroup".}
	\label{fig:energy_dos}
\end{figure*}

dRNN is formulated in the fashion that the gates are modulated by the weighted combinations of several orders of DoS.
While an individual order of DoS is able to model a certain degree of dynamical structures, the sum of all the orders of DoS could distort the detected salient motion patterns. Figure \ref{fig:energy_dos} illustrates the energy curves of the 0th-, 1st-, and 2nd-orders of DoS over an example of sequence for the activity "RunInGroup". The local maxima indicate high energy landmarks corresponding to the salient motion frames at different levels. While each order of DoS enables the LSTM unit to model the dynamics of local saliency at a certain level, the weighted sum of different orders of DoS may risk misaligning salient motion and result in distorted motion patterns.
This inspires us to control the LSTM gates using individual orders of the state derivatives. 

As is generally accepted, RNNs are inherently deep in time because the current hidden state is a function of all previous hidden states. By questioning whether RNNs could also benefit from depth in space, just as feed-forward layers which are stacked in conventional deep networks, Graves \textit{et al}. \cite{graves2013speech} introduced Deep Recurrent Neural Networks, also known as stacked LSTMs. Stacked LSTMs have shown superiority over the traditional LSTM in modeling complex sequences and have been used in various types of applications.
Inspired by Deep Recurrent Neural Network, we are motivated to explore whether the stacked deep layers in space could naturally reveal the saliency dynamics over time, thus avoiding the misaligned DoS in different orders.

To this end, we propose to stack multiple levels of LSTM cells with increasing orders of DoS.
The proposed model progressively builds up the ability of LSTM gates to detect salient dynamic patterns with deeper memory layers modeling higher orders of DoS. The proposed model is thus termed deep differential Recurrent Neural Network ($d^2$RNN). The $d^2$RNN differs from conventional stacked LSTMs in that stacked LSTMs use homogeneous LSTM layers while $d^2$RNN uses heterogeneous ones. In this way, $d^2$RNN is not only capable of modeling more complex dynamical patterns, but also enables a hierarchy of DoS saliency in deep layers to model the spatio-temporal dynamics over time.




We demonstrate that $d^2$RNN can outperform the state-of-the-art methods on two publicly available human activity datasets: NUS-HGA \cite{ni2009recognizing} and Violent-Flows \cite{hassner2012violent}. Specifically, $d^2$RNN outperforms the existing LSTM, dRNN, and stacked LSTM models, consistently achieving better performance in detecting human activities in sequences. In addition, we compared with the other non-LSTM algorithms, where $d^2$RNN model also reached competitive performance.

The remainder of this paper is organized as follows. In the next section, we briefly review several related works. The background and details of dRNN are reviewed in Section \ref{sec:background}. Section \ref{sec:model} presents the proposed deep differential RNN model. The experimental results are presented in Section \ref{sec:exp}. Finally, we offer our conclusion and discuss the future work in Section \ref{sec:conclusion}.

\section{Related Work}
\label{sec:related}

\subsection{Variants of Long Short-Term Memory}
Due to the exponential decay, traditional RNNs are limited in learning long-term sequences. Hochreiter \textit{et al}. \cite{hochreiter1997long} designed Long Short-Term Memory (LSTM) to exploit the long-range dependency. As LSTM shows superiority in modeling time-series data, it is widely used to various kinds of sequential processing tasks and several LSTM variants were proposed to improve the architecture of standard LSTM. S-LSTM \cite{zhu2015long} is an LSTM network with tree structures. The hierarchical structure of S-LSTM aims to mitigate the gradient vanishing problem and model more complicated input than LSTM. Bidirectional LSTM \cite{schuster1997bidirectional} captures both future and past context of the input sequence. Multidimensional LSTM (MDLSTM) \cite{DBLP:journals/corr/abs-0705-2011} uses interconnection from previous state of cell to extend the memory of LSTM along every $N$-dimension. The MDLSTM receives inputs in an $N$-dimensional arrangement, thus can model multidimensional sequences. MDLSTM model becomes unstable with the growth of the grid size and LSTM depth in space. Grid LSTM \cite{kalchbrenner2015grid} provides a solution by altering the computation of output memory vectors. 

Even though the above variants of LSTMs show superiority in some aspect, they do not consider the salient spatio-temporal dynamics which can be modeled by information gain of internal memory states. This inspires the use of Derivative of States (DoS) in differential Recurrent Neural Networks (dRNN) \cite{veeriah2015differential}. Unfortunately, dRNN uses the weighted combination of DoS to modulate the LSTM gates, which could distort the detected salient motion patterns. We are motivated to control the gates using individual orders of DoS.

Stacked LSTMs \cite{graves2013speech} borrow the idea of depth in ANNs and stack hidden layers with LSTM cells in space to increase the network capacity. However, the homogeneous layers of stacked LSTMs limit its ability to model discriminative spatio-temporal structures. We are motived to explore a hierarchy of DoS saliency in deep layers.

\begin{figure*}
	\centering
		\includegraphics[width=1.0\linewidth]{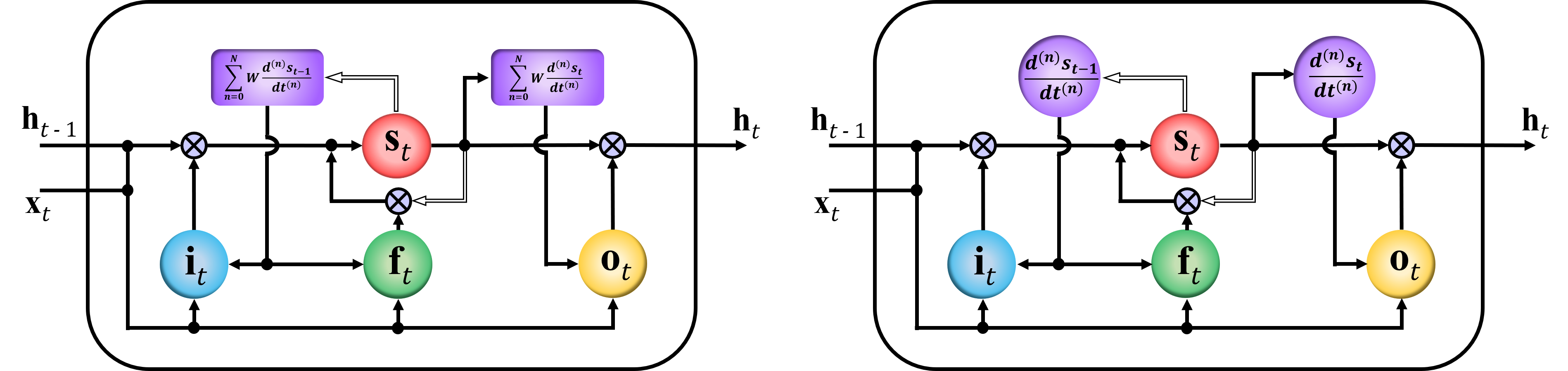}
	\caption{Architectures of dRNN cell (left) and nth-order (layer $n+1$) of $d^2$RNN cell (right) at time $t$. \textbf{Best viewed in color}. }
	\label{fig:drnn_d2rnn}
\end{figure*}

\subsection{Activity Recognition}

Human activity recognition includes sub-problems of individual human action recognition and multi-person activity recognition. In this paper, we focus on the more challenging problem of multi-person activity recognition. Multi-person activity recognition is further divided into group activity recognition and crowd analysis.


Most existing approaches of group activity recognition are based on motion trajectories of group participants. Ni \emph{et~al.} \cite{ni2009recognizing} applied motion trajectory segments as inputs and used digital filters' frequency responses to represent the motion information. Zhu \emph{et~al.} \cite{zhu2011generative} considered motion trajectory as a dynamic system and used the Markov stationary distribution to acquire local appearance features as a descriptor of group action. Chu \emph{et~al.} \cite{chu2012new} designed an algorithm to model the trajectories as series of heat sources to create a heat map for representing group actions.
Cho \emph{et~al.} \cite{cho2015group} addressed the problem by using group interaction zones to detect meaningful groups to handle noisy information.
Cheng \emph{et~al.} \cite{cheng2014recognizing} proposed a layered model of human group action and represented activity patterns with both motion and appearance information. Their performance on NUS-HGA achieved an accuracy of 96.20\%. 
Zhuang \textit{et al.} \cite{zhuang2017group} used a combination of Deep VGG network \cite{simonyan2014very} and stacked LSTMs. Their model complexity is high and has a large chance of overfitting. In this case, their model is trained on augmented data thus cannot be fairly compared with other methods.

Recent methods for crowd scene understanding mostly analyze crowd activities based on motion features extracted from trajectories/tracklets of objects \cite{shao2014scene, hassner2012violent, su2016crowd, mousavi2015analyzing}. Marsden et al. \cite{marsden2016holistic} studied scene-level holistic features using tracklets to solve real-time crowd behavior anomaly detection problem. \cite{marsden2016holistic} holds the state-of-the-art performance for the Violent-Flows dataset. Su et al. \cite{su2016crowd} used tracklet-based features and explored Coherent LSTM to model the nonlinear characteristics and spatio-temporal motion patterns in crowd behaviors. The trajectory/tracklet feature contains more semantic information, but the accuracy of trajectories/tracklets dictates the performance of crowd scene analysis. In extremely crowded areas, tracking algorithms could fail and generate inaccurate trajectories. The general-purpose $d^2$RNN does not require such input, holding the potential for more sequence-related applications.

\section{Background}
\label{sec:background}

In this section, we briefly review Recurrent Neural Network (RNN) as well as differential Recurrent Neural Network (dRNN). Readers who are familiar with them might skip to the next section directly.

\subsection{Recurrent Neural Networks}

Traditional Recurrent Neural Networks model the dynamics of an input sequence of frames $\{\textbf{x}_t \in \mathds{R}^{m} | t = 1, 2, ..., T\}$ through a sequence of hidden states
$\{\textbf{h}_t \in \mathds{R}^{n} | t = 1, 2, ..., T\}$, thereby learning the spatio-temporal structure of the input sequence. For instance, a classical RNN model uses the following recurrent equation
\begin{equation}
	\textbf{h}_t = \tanh(\textbf{W}_{hh}\textbf{h}_{t-1} + \textbf{W}_{hx}\textbf{x}_{t} + \textbf{b}_h),
\end{equation}
to model the hidden state $\textbf{h}_t$ at time $t$ by combining the information from the current input $\textbf{x}_t$ and the past hidden state $\textbf{h}_{t-1}$. The hyperbolic tangent $\tanh(\cdot)$ in the above equation is an activation function with range [-1, 1], $\textbf{W}_{hh}$ and $\textbf{W}_{hx}$ are two mapping matrices to the hidden states, and $\textbf{b}_h$ is the bias vector.

The hidden states will then be mapped to an output sequence $\{\textbf{y}_t \in \mathds{R}^{k} | t = 1, 2, ..., T\}$ as
\begin{equation}{\label{eq:out__}}
	\textbf{y}_t = \tanh(\textbf{W}_{yh}\textbf{h}_{t} + \textbf{b}_y),
\end{equation}
where each $\textbf{y}_t$ represents a 1-of-$k$ encoding of the confidence scores on $k$ classes of human activities. This output can then be transformed to a vector of probabilities $\textbf{p}_t$ by the softmax function as
\begin{equation}{\label{eq:softmax}}
	p_{t,c} = \frac{\exp(y_{t,c})}{\sum_{l=1}^k\exp(y_{t,l})},
\end{equation}
where each entry $p_{t,c}$ is the probability of frame $t$ belonging to class $c \in \{1,...,k\}$.

\subsection{Differential Recurrent Neural Networks}

Due to exponential decay in retaining the context information from video frames, traditional RNNs are limited in learning the long-term representation of sequences. Hochreiter \textit{et al.} \cite{hochreiter1997long} designed Long Short-Term Memory (LSTM) to exploit the long-range dependency. 


Although traditional LSTM neural network is capable of processing complex sequential information, it might fail to capture the salient dynamic patterns because the gate units do not \textit{explicitly} consider the impact of dynamic structures present in input sequences. This makes the conventional LSTM model inadequate to learn the evolution of action states. Veeriah \textit{et al.} \cite{veeriah2015differential} introduced the Derivate of States (DoS) for dRNN, which can explicitly model spatio-temporal structure and learn salient motion patterns within. Replacing internal state with the DoS in the gate units, dRNN has the following updated equations:

 

(i) Input gate $\textbf{i}_{t}$ regulates the degree to which the input information would enter the memory cell to affect the internal state $\textbf{s}_t$ at time $t$. The activation of the gate has the following recurrent form:
\[\textbf{i}_{t} = \sigma(\sum_{n=0}^{N}\textbf{W}^{(n)}_{id}\frac{d^{(n)}\textbf{s}_{t-1}}{dt^{(n)}} + \textbf{W}_{ih}\textbf{h}_{t-1} + \textbf{W}_{ix}\textbf{x}_{t} + \textbf{b}_i),\]
where the sigmoid $\sigma(\cdot)$ is an activation function in the range [0,1], with 0 meaning the gate is closed and 1 meaning the gate is completely open; $\textbf{W}_{i*}$ are the mapping matrices and $\textbf{b}_i$ is the bias vector.

(ii) Forget gate $\textbf{f}_t$ modulates the previous state $\textbf{s}_{t-1}$ to control its contribution to the current state. It is defined as
\[\textbf{f}_{t} = \sigma(\sum_{n=0}^{N}\textbf{W}^{(n)}_{fd}\frac{d^{(n)}\textbf{s}_{t-1}}{dt^{(n)}} + \textbf{W}_{fh}\textbf{h}_{t-1} + \textbf{W}_{fx}\textbf{x}_{t} + \textbf{b}_f),\] with the mapping matrices $\textbf{W}_{f*}$ and the bias vector $\textbf{b}_f$.

With the input and forget gate units, the internal state $\textbf{s}_t$ of each memory cell can be updated as below:
\begin{equation}\label{eq:inter_state}
 \textbf{s}_t = \textbf{f}_t \otimes \textbf{s}_{t-1} + \textbf{i}_t \otimes \tanh(\textbf{W}_{sh}\textbf{h}_{t-1} + \textbf{W}_{sx}\textbf{x}_t + \textbf{b}_s),  
\end{equation}
where $\otimes$ stands for element-wise product. 

(iii) Output gate $\textbf{o}_t$ gates the information output from a memory cell which would influence the future states of LSTM cells. It is defined as
\[\textbf{o}_{t} = \sigma(\sum_{n=0}^{N}\textbf{W}^{(n)}_{od}\frac{d^{(n)}\textbf{s}_{t}}{dt^{(n)}} + \textbf{W}_{oh}\textbf{h}_{t-1} + \textbf{W}_{ox}\textbf{x}_{t} + \textbf{b}_o).\]
Then the hidden state of a memory cell is output as 
\begin{equation}\label{eq:out}
\textbf{h}_t = \textbf{o}_t \otimes \tanh(\textbf{s}_t).
\end{equation}
By iteratively applying Eq. \ref{eq:inter_state} and Eq. \ref{eq:out}, dRNN updates the internal state $\textbf{s}_t$ and the hidden state $\textbf{h}_t$ over time. In the process, the input gate, forget gate, and output gate play an important role in controlling the information entering and leaving the memory cell. 

\section{The proposed model}
\label{sec:model}

\begin{figure*}
	\centering
		\includegraphics[width=1.0\linewidth]{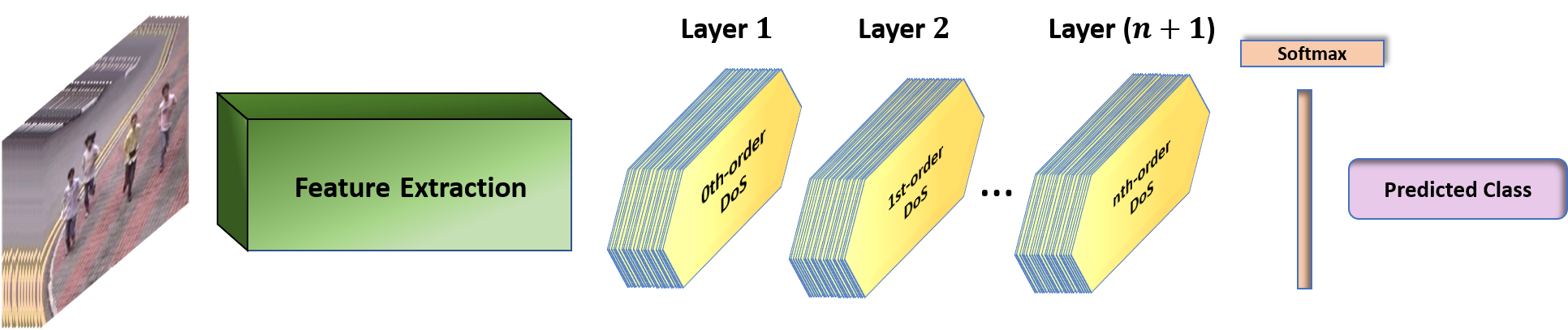}
	\caption{An illustration of our framework for video-level activity recognition. \textbf{Best viewed in color}. }
	\label{fig:pipeline}
\end{figure*}

Given an activity recognition task, not all video frames contain salient patterns to discriminate between different classes of activities. 
dRNN tries to detect and integrate the salient spatio-temporal sequences via the state derivative. As the internal state contains the memory of the previous input sequences, the state derivative explicitly models the change in information gain and considers the impact of dynamic structures.
Thus, the state derivative tends to be effective in recognizing actions. 

As mentioned in \cite{greff2017lstm}, LSTM gates serve as the most crucial elements of LSTM. 
dRNN formulates the input, forget, and output gates using the combination of different orders of DoS. To be more specific, 0th-order DoS, which is the same as conventional LSTM internal cell, models the locality of memory change; 1st-order DoS denotes the velocity of change in information gain; and 2nd-order DoS describes the acceleration of memory change, etc. While each individual order of DoS is effective in capturing a certain level of salient spatio-temporal sequences, the sum of all the orders of DoS could distort the detected salient motion patterns and result in less effective modulation of those gates. 

In this paper, we propose to modulate the LSTM gates via individual orders of DoS.
Inspired by \cite{graves2013speech}, we stack multiple levels of LSTM cells with increasing orders of DoS. To be more specific, layer 1 of $d^2$RNN uses 0th-order DoS, which resembles the conventional LSTM cell; layer 2 uses LSTM cell with 1st-order DoS; and layer 3 uses LSTM cell with 2nd-order DoS, etc. Since we are integrating the ideas of DoS from dRNN and deep stacked layers from deep RNN, our proposed model is termed deep differential Recurrent Neural Network ($d^2$RNN). 
Within each layer of $d^2$RNN, our model learns the change in information gain with individual order of DoS. With deeper layers of $d^2$RNN cell, our model learns higher-degree and more complex dynamical patterns.

Figure \ref{fig:drnn_d2rnn} illustrates the LSTM unit in layer $(n+1)$ of the proposed $d^2$RNN model. Hollow lines indicate the information flow of $\textbf{s}_{t-1}$. Formally, we have the following recurrent equations to control the LSTM gates in layer $(n+1)$ of $d^2$RNN.

(i) Input gate:
\begin{equation}\label{eq:input}
	\textbf{i}_{t} = \sigma(\textbf{W}^{(n)}_{id}\frac{d^{(n)}\textbf{s}_{t-1}}{dt^{(n)}} + \textbf{W}_{ih}\textbf{h}_{t-1} + \textbf{W}_{ix}\textbf{x}_{t} + \textbf{b}_i),
\end{equation}

(ii) Forget gate:
\begin{equation}\label{eq:forget}
	\textbf{f}_{t} = \sigma(\textbf{W}^{(n)}_{fd}\frac{d^{(n)}\textbf{s}_{t-1}}{dt^{(n)}}  + \textbf{W}_{ih}\textbf{h}_{t-1} + \textbf{W}_{fx}\textbf{x}_{t} + \textbf{b}_f),
\end{equation}

(iii) Output gate:
\begin{equation}\label{eq:output}
	\textbf{o}_{t} = \sigma(\textbf{W}^{(n)}_{od}\frac{d^{(n)}\textbf{s}_{t}}{dt^{(n)}}  + \textbf{W}_{ih}\textbf{h}_{t-1} + \textbf{W}_{ox}\textbf{x}_{t} + \textbf{b}_o),
\end{equation}



\subsection{Discretized Model}

Since $d^2$RNN model is defined in the discrete-time domain, the 1st-order derivative $\frac{d\textbf{s}_t}{dt}$, as the velocity of information change, can be discretized as the difference of states:
\begin{equation}{\label{eq:vel}}
	\textbf{v}_t \triangleq \frac{d\textbf{s}_t}{dt} \doteq \textbf{s}_t - \textbf{s}_{t-1},
\end{equation}
for simplicity \cite{epperson2013introduction}.

Similarly, we consider the 2nd-order of DoS as the acceleration of information change. It can be discretized as:
\begin{equation}{\label{eq:acc}}
	\textbf{a}_t \triangleq \frac{d^2\textbf{s}_t}{dt^2} \doteq \textbf{v}_t - \textbf{v}_{t-1} = \textbf{s}_t - 2\textbf{s}_{t-1} + \textbf{s}_{t-2}.
\end{equation}

In this paper, we only consider the first two orders of DoS. Higher orders can be derived in a similar way.

\subsection{Algorithm and Learning}


With the above recurrent equations, the $d^2$RNN model proceeds with the following procedures starting in layer 1 ($n = 0$) at time step $t$:
\begin{itemize}
	\item Compute input gate activation $\textbf{i}_t$ and forget gate activation $\textbf{f}_t$ by Eq. (\ref{eq:input}) and Eq. (\ref{eq:forget});
	\item Update state $\textbf{s}_t$ with $\textbf{i}_t$ and $\textbf{f}_t$ by Eq. (\ref{eq:inter_state});
    \item Compute discretized DoS $\frac{d^{(n)}\textbf{s}_{t}}{dt^{(n)}}$;
	\item Compute output gate $\textbf{o}_t$ by Eq. (\ref{eq:output});
	\item Output $\textbf{h}_t$ gated by $\textbf{o}_t$ from memory cell by Eq. (\ref{eq:out});
    \item If there exists a deeper layer in $d^2$RNN, set $n = n + 1$ and $\textbf{x}_t = \textbf{h}_t$ for the following layer and repeat the above steps;
    
\end{itemize}



For a frame-by-frame prediction task, we output the label $\textbf{p}_t$ by applying the softmax to $\textbf{h}_t$ using Eq. (\ref{eq:out__}) and (\ref{eq:softmax}).
To learn the model parameters of $d^2$RNN, we define a loss function to measure the deviation between the target class $c_t$ and $\textbf{p}_t$ at time $t$:
\[\ell(\textbf{p}_t,c_t) = -\log p_{t,{c_t}}.\]
Then, we can minimize the cumulative loss over the sequence:
\[\sum_{t=1}^{T}\ell(\textbf{p}_t,c_t).\]

For an activity recognition task, the label of activity is often given at the video level.
Since LSTMs have the ability to memorize the content of an entire sequence, the last memory cell of LSTMs ought to contain all the necessary information for recognizing the activity. The sequence level class probability $\textbf{p}$ is generated by computing the output of $d^2$RNN with Eq. (\ref{eq:out__}) and applying the softmax function with Eq. (\ref{eq:softmax}). For a given training label $c$, the $d^2$RNN can be trained by minimizing the loss function below, i.e.
\[\ell(\textbf{p},c) = -\log p_{c}.\]

The loss function can be minimized by Back Propagation Through Time (BPTT) \cite{cuellar2007application}, which unfolds an LSTM model over several time steps and then runs the back propagation algorithm to train the model. To prevent back-propagated errors from decaying or exploding exponentially, we use truncated BPTT according to Hochreiter \emph{et~al.} \cite{hochreiter1997long} to learn the model parameters. Specifically, in our model, errors are not allowed to re-enter the memory cell once they leave it through the DoS nodes.

  Formally, we assume the following truncated derivatives of gate activations:
\[ \frac{\partial \textbf{i}_t}{\partial \textbf{v}_{t-1}} \circeq 0, \frac{\partial \textbf{f}_t}{\partial \textbf{v}_{t-1}} \circeq 0, \frac{\partial \textbf{o}_t}{\partial \textbf{v}_{t}} \circeq 0,\]
and
\[ \frac{\partial \textbf{i}_t}{\partial \textbf{a}_{t-1}} \circeq 0, \frac{\partial \textbf{f}_t}{\partial \textbf{a}_{t-1}} \circeq 0, \frac{\partial \textbf{o}_t}{\partial \textbf{a}_{t}} \circeq 0,\]
where $\circeq$ stands for the truncated derivatives.

\section{Experimental Results}
\label{sec:exp}

We compare the performance of the proposed method with state-of-the-art LSTM and non-LSTM methods present in existing literature on human activity datasets.

\subsection{Datasets and Feature Extraction}
The proposed method is evaluated on two publicly available human activity datasets: NUS-HGA \cite{ni2009recognizing} and Violent-Flow \cite{hassner2012violent}.

We choose the NUS-HGA dataset as it is a well-collected benchmark dataset for evaluating activity recognition techniques. 
The NUS-HGA dataset includes 476 video clips covering six group activity classes: Fight, Gather, Ignore, RunInGroup, StandTalk and, WalkInGroup. Each instance involves 4-8 persons. The sequences are captured over different backgrounds with a static camera recording 25 frames per second. Each video clip has a resolution of 720 $\times$ 576 and lasts around 10 seconds.

The Violent-Flows (VF) dataset is a real-world video footage of crowd violence, along with standard benchmark protocols designed for violent/non-violent classification. The Violent-Flows dataset includes 246 real-world videos downloaded from YouTube. The shortest clip duration is 1.04 seconds, the longest slip is 6.52 seconds, and the average length of the video is 3.6 seconds.


We are using densely sampled HOG3D features \cite{klaser2008spatio} to represent each frame of video sequences from the NUS-HGA and Violent-Flows datasets. Specifically, we uniformly divide the 3D video volumes into a dense grid, and extract the descriptors from each cell of the grid. The parameters for HOG3D are the same as the one used in \cite{klaser2008spatio}. 
The standard dense sampling parameters for extracting HOG3D features can be found on the author's webpage.
All the videos are resized into $160 \times 120$. The size of the descriptor is 300 per cell of grid, and there are 58 such cells in each frame, yielding a 17,400 dimensional feature vector per frame. To construct a compact input into $d^2$RNN model, Principal Component Analysis (PCA) is then applied to reduce the feature dimension. After PCA dimension reduction, NUS-HGA has a feature dimension of 300, and Violent-Flows has a feature dimension of 500. Both of them are retaining 90\% of energy among the principal components. For the sake of fair comparison, we use the same features as input for other LSTM models too.

\subsection{Architecture and Training}

The architecture of the $d^2$RNN models trained on the above datasets are shown in Table \ref{tab:archit}. 
We keep the state units the same for all the LSTM layers in $d^2$RNN. 
For the sake of fair comparison, we adopt the same architecture for stacked LSTMs models. For dRNN model, we keep the same number of state units as stacked LSTMs and $d^2$RNN.
We can see that the number of memory cell units is smaller than the input units on both datasets. This can be interpreted as follows. The sequence of a human activity video often forms a continuous pattern embedded in a low-dimensional manifold of the input space. Thus, a lower-dimension state space is sufficient to capture the dynamics of such patterns. The number of output units corresponds to the number of classes in the datasets.

\begin{table}
	\centering
		\begin{tabular}{ c | c | c }
		\hline
			 & \textbf{NUS-HGA} & \textbf{Violent-Flows} \\ 
		\hline
			Input Units & 300 & 500 \\
			State Units & 200 & 300\\
			Output Units & 6 & 2 \\
		\hline	
		\end{tabular}
	\caption{Architectures of the $d^2$RNN model used on the NUS-HGA and Violent-Flows datasets.}
	\label{tab:archit}
\end{table}

\begin{figure}
	\centering
		\includegraphics[width=1.0\linewidth]{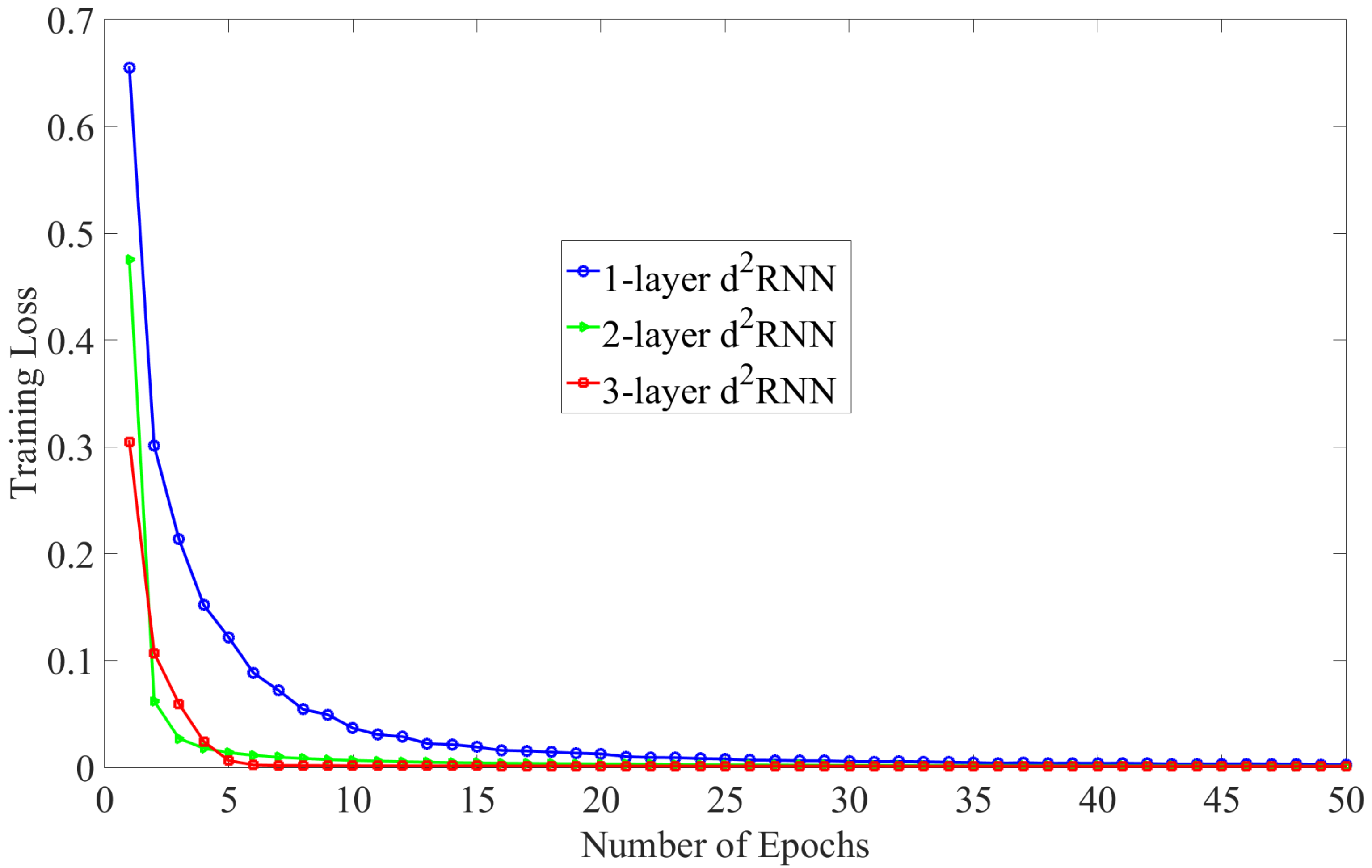}
	\caption{Training loss curves of different layers of $d^2$RNN over number of epochs on the NUS-HGA dataset.}
	\label{fig:loss}
\end{figure}

We plot the learning curves for training the $d^2$RNN models on the NUS-HGA dataset in Figure \ref{fig:loss}. $(n+1)$-layer $d^2$RNN refers to the model of $(n+1)$ layers with DoS starting from 0th-order to nth-order. The learning rate of BPTT algorithm is set to 0.0001. The figure shows that the objective loss continuously decreases over 50 epochs. Deep layers of $d^2$RNN converge faster due to larger model complexity.

\subsection{Results on the NUS-HGA Dataset}

\begin{figure*}
	\centering
		\includegraphics[width=1.0\linewidth]{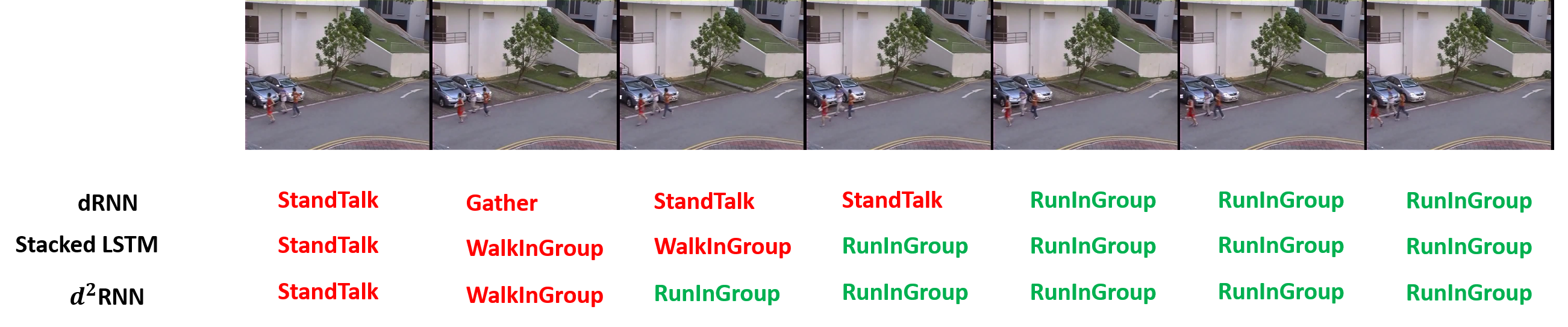}
	\caption{Frame-by-frame prediction of activity category over time. \textbf{Best viewed in color}.}
	\label{fig:frame}
\end{figure*}

There are several different evaluation protocols used on the NUS-HGA dataset in literature, which can lead to fairly large differences in performance across different experiment protocols. 

First to evaluate the performances of the proposed $d^2$RNN vs. other LSTM models, we perform five-fold cross validation. 
This experiment set-up increases the challenge for the task compared to the one using Monte-Carlo cross validation. For the NUS-HGA dataset, the activity videos are produced by chopping longer sequences into shorter ones. 
Due to random sampling, Monte-Carlo makes the task easier by putting similar video instances to both training and testing set.
Five-fold cross-validation, however, dramatically increases the difficulty since training and testing examples usually have high in-class variations regarding background, view-angle, lightning, and activity participants.

\begin{figure}
	\centering
		\includegraphics[width=1.0\linewidth]{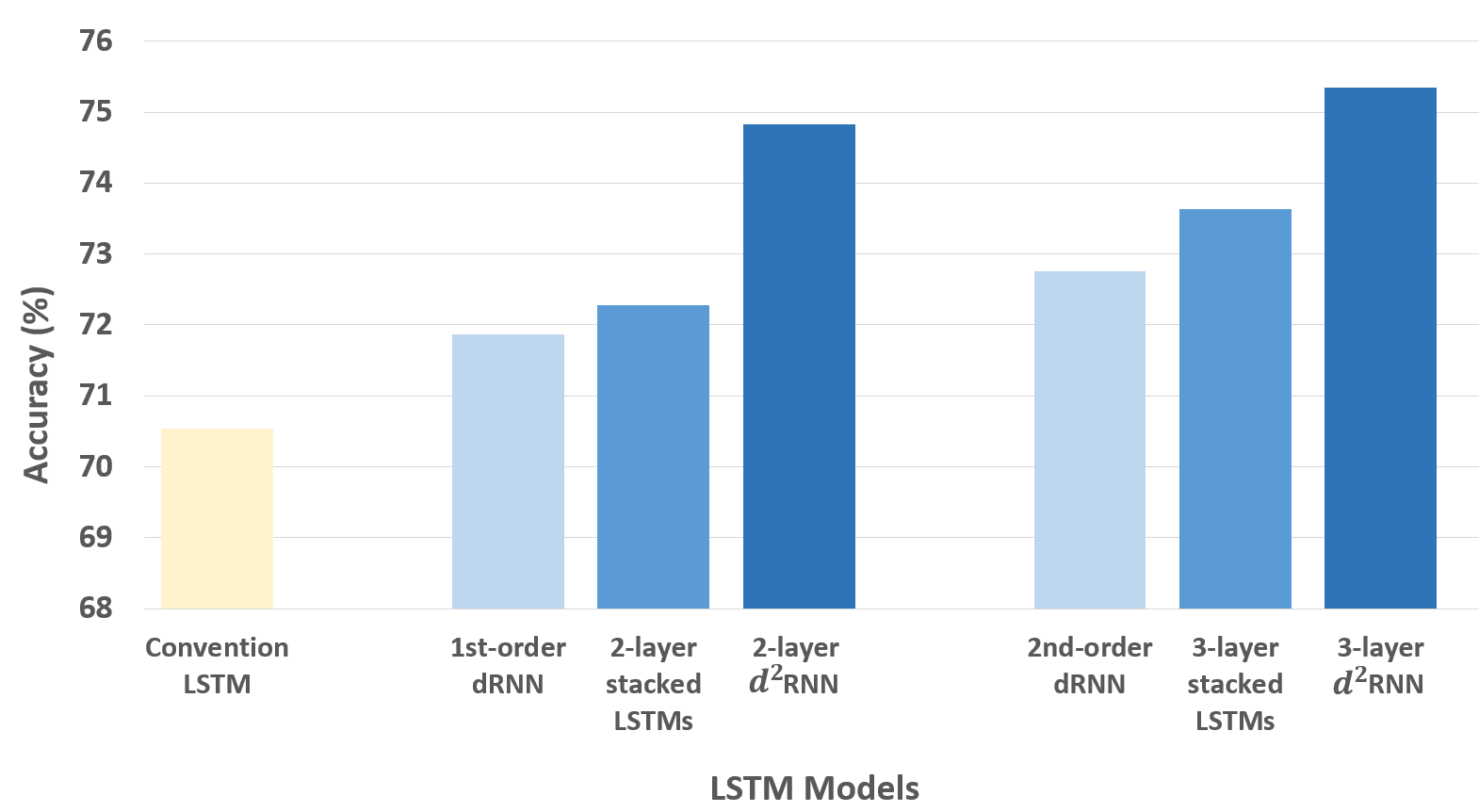}
	\caption{Comparison of the proposed $d^2$RNN model with other LSTM models on the NUS-HGA dataset.}
	\label{fig:nus_lstm}
\end{figure}
We show the performance comparison of LSTM models in Figure \ref{fig:nus_lstm}. For the dRNN and $d^2$RNN models, we report the accuracy up to the 2nd-order of DoS. For stacked LSTMs, we report the accuracy up to 3 layers, which is the same highest layer as $d^2$RNN. All the LSTM models use the same HOG3D feature presented above. The conventional LSTM yields the lowest performance compared to other LSTM models. This is because conventional LSTM uses neither DoS or deep layers to capture motion saliency presented in given video frames.

Generally, higher orders of DoS generate better performance for both dRNN and $d^2$RNN. However, the performance increase for dRNN is smaller than $d^2$RNN. This can be explained that for dRNN, the combination of all the orders of DoS distorts the detected motion patterns. $d^2$RNN, on the other hand, uses individual orders of DoS on each layer, and can preserve and align the salient dynamic structures. 

It can also be seen that $d^2$RNN outperforms stacked LSTMs given the same number of deep layers, which demonstrates the advantage of heterogeneous LSTM layers used in $d^2$RNN over the homogeneous ones in stacked LSTMs. To be more specific, the higher orders of DoS from $d^2$RNN detects salient spatio-temporal structures which cannot be captured by conventional LSTM layers used in stacked LSTMs. To this end, the above analysis shows that $d^2$RNN not only is capable of modeling more complex dynamical patterns, but also enables a hierarchy of DoS saliency in deep layers to model the spatio-temporal dynamics over time.

Although deeper layers or higher orders of $d^2$RNN might improve the accuracy further, we do not report the result since it becomes trivial by simply adding more deep layers with higher-order DoS. The improved performance, however, might not compensate for the increased computational cost. Moreover, with an increased number of deep layers modeling higher orders of DoS, a larger number of model parameters would have to be learned with the limited training examples. This tends to cause overfitting problem, making the performance stop improving or even begin to degenerate. Therefore, for most of practical applications, the 3-layer setup for $d^2$RNN should be sufficient. 
More applications of deep architectures of RNNs can be found in \cite{irsoy2014opinion, sutskever2014sequence}.

In order to compare $d^2$RNN with other non-LSTM state-of-the-art methods, we follow \cite{cheng2014recognizing} and evaluate our method via Monte-Carlo Cross Validation. To be more specific, we randomly select 80\% of the examples from each class of the dataset to form the training set, and then assign the rest to the test set. The performance is reported by the average accuracy across five such trials.

\begin{table}
	\centering
		\begin{tabular}{ c | c  }
		\hline
			\textbf{Methods} & \textbf{Accuracy (\%)} \\ 
		\hline
		    Ni \emph{et~al.} \cite{ni2009recognizing} & 73.50 \\
			Zhu \emph{et~al.} \cite{zhu2011generative} & 87.00 \\
			Cho \emph{et~al.} \cite{cho2015group} & 96.03\\
			Cheng \emph{et~al.} \cite{cheng2014recognizing} (MF) & 93.20\\
			Cheng \emph{et~al.} \cite{cheng2014recognizing} (MAF) & 96.20\\
            3-layer $d^2$RNN & \textbf{98.24} \\
		\hline
		\end{tabular}
	\caption{Performance comparison on the NUS-HGA dataset. MF indicates motion feature fusion and MAF indicates motion and appearance feature fusion.}
	\label{tab:HGA}
\end{table}

\begin{figure}
	\centering
		\includegraphics[width=1.0\linewidth]{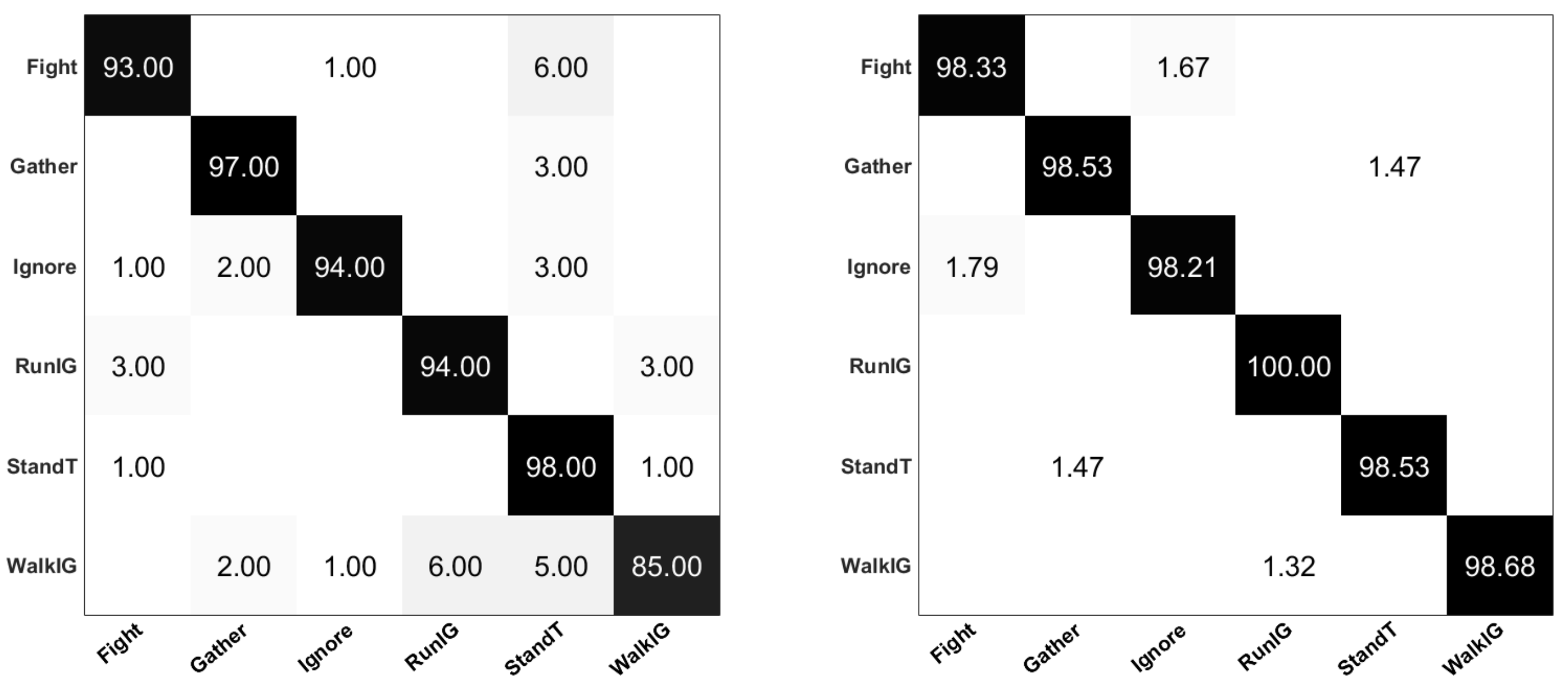}
	\caption{Confusion Matrices obtained by Cheng \textit{et al}. \cite{cheng2014recognizing} (left) and 3-layer $d^2$RNN (right) on the NUS-HGA dataset.}
	\label{fig:msr_cf}
\end{figure}

We compare $d^2$RNN model with other non-LSTM state-of-the-art algorithms in Table \ref{tab:HGA}. In addition, Figure \ref{fig:msr_cf} shows the confusion matrices obtained by \cite{cheng2014recognizing} and our proposed method.
$d^2$RNN model generally achieves better performance than the other non-LSTM methods. Traditional solutions for group activity algorithms need human supervision to acquire accurate human object trajectories from the videos. According to \cite{cheng2014recognizing}, they acquired human bounding boxes using existing tracking tools, which requires manual annotation for bounding box initialization. This constraint prevents their method from being used in automatic or real-time applications. On the contrary, $d^2$RNN models can outperform these traditional methods without the aid of manual operation, enabling broader applications for group behavior analysis.
In addition, since traditional models rely on such special assumptions, they might not be applicable to other types of sequences which do not satisfy these assumptions. 
In contrast, the proposed $d^2$RNN model is a general-purpose model, not being tailored to any specific type of sequences, holding the potential for other sequence-related applications. 

Zhuang \cite{zhuang2017group} reported an accuracy of 99.25\% on NUS-HGA dataset. 
It is worth noting that they used a combination of Deep VGG network \cite{simonyan2014very} with stacked LSTMs, where the Deep VGG network plays a crucial role in reaching such performance. In addition, to avoid overfitting for the complex model, they applied data augmentation on the training set to increase the diversity of training sequences. While in our experiments, no data augmentation technique is used.

In order to better understand the disadvantage of the combination of different orders of DoS in dRNN model, we perform the following experimental study. First, we construct a variant of LSTM using individual orders of DoS and then treat them as separate models. We call these models "1st-order LSTM" and "2nd-order LSTM". Then we use AdaBoost algorithm \cite{freund1997decision} to ensemble the above models, which we call Ensemble RNN (eRNN). To be more specific, 1st-order eRNN ensembles 0th- and 1st-order LSTMs; 2nd-order eRNN ensembles 0th-, 1st-, and 2nd-order LSTMs. By doing so, we also intend to study whether each individual order of DoS is good at modeling a certain level of motion saliency. 

\begin{figure}
	\centering
		\includegraphics[width=1.0\linewidth]{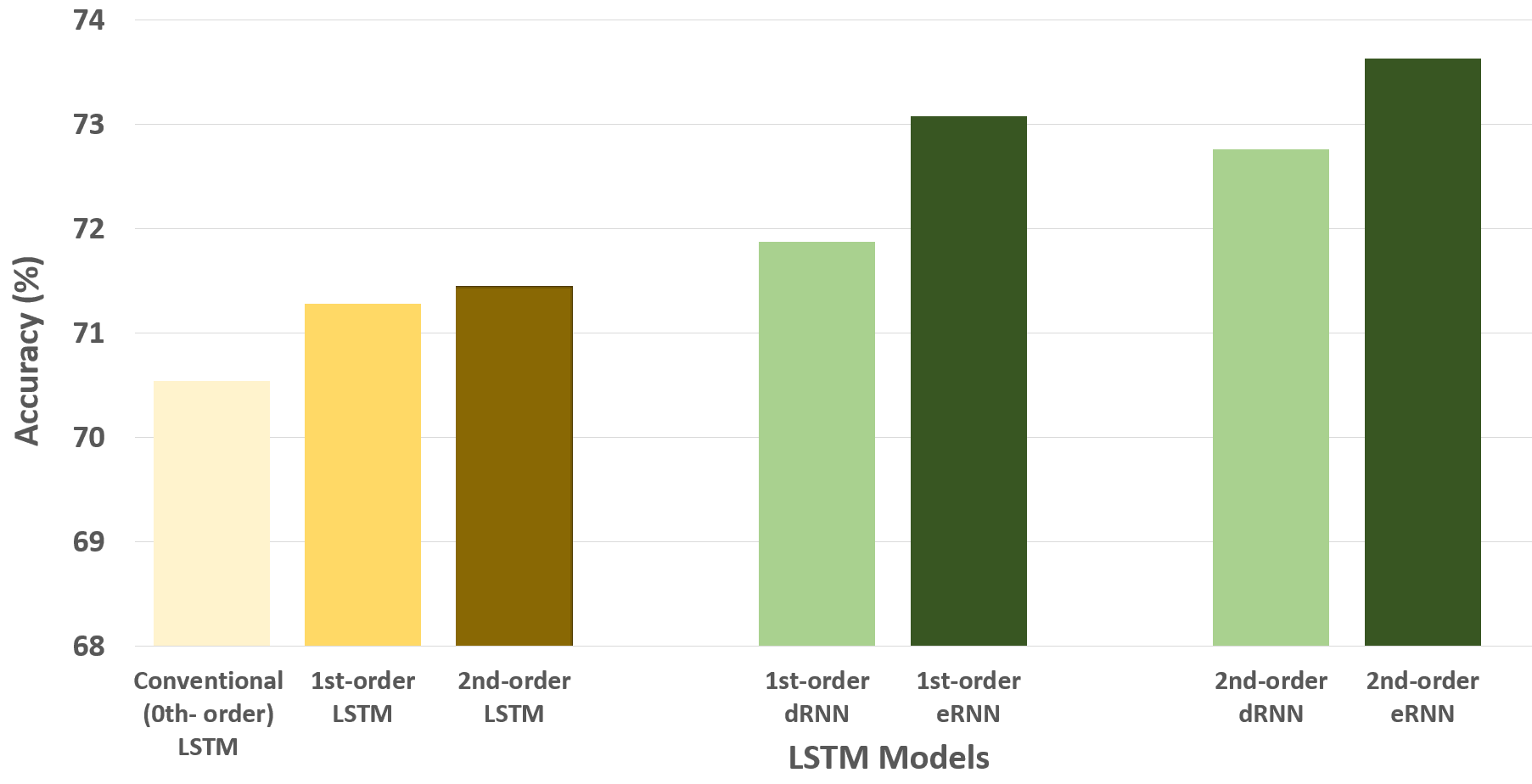}
	\caption{Evaluation of individual vs. combination of DoS.}
	\label{fig:drnn_ensemble}
\end{figure}

In Figure \ref{fig:drnn_ensemble}, the leftmost three bars are performances of conventional LSTM, 1st-order LSTM, and 2nd-order LSTM. LSTMs with higher orders of individual DoS gain slightly better performances. 
On the other hand, their ensemble models, which are eRNNs, achieve substantially better results.
This shows that each individual order of DoS indeed can detect a certain level of motion saliency and contributes to their ensemble model. While comparing the same order of dRNN with eRNN, we find out that eRNN consistently achieves higher results. This demonstrates that it is suboptimal to combine different orders of DoS within the LSTM gates and the sum of all the orders of DoS would distort the detected motion patterns.

\subsection{Results on the Violent-Flows Dataset}

To evaluate our method on the Violent-Flows dataset, we follow the standard 5-fold cross-validation protocol in \cite{hassner2012violent} and report the results in terms of mean accuracy. 


\begin{figure}
	\centering
		\includegraphics[width=1.0\linewidth]{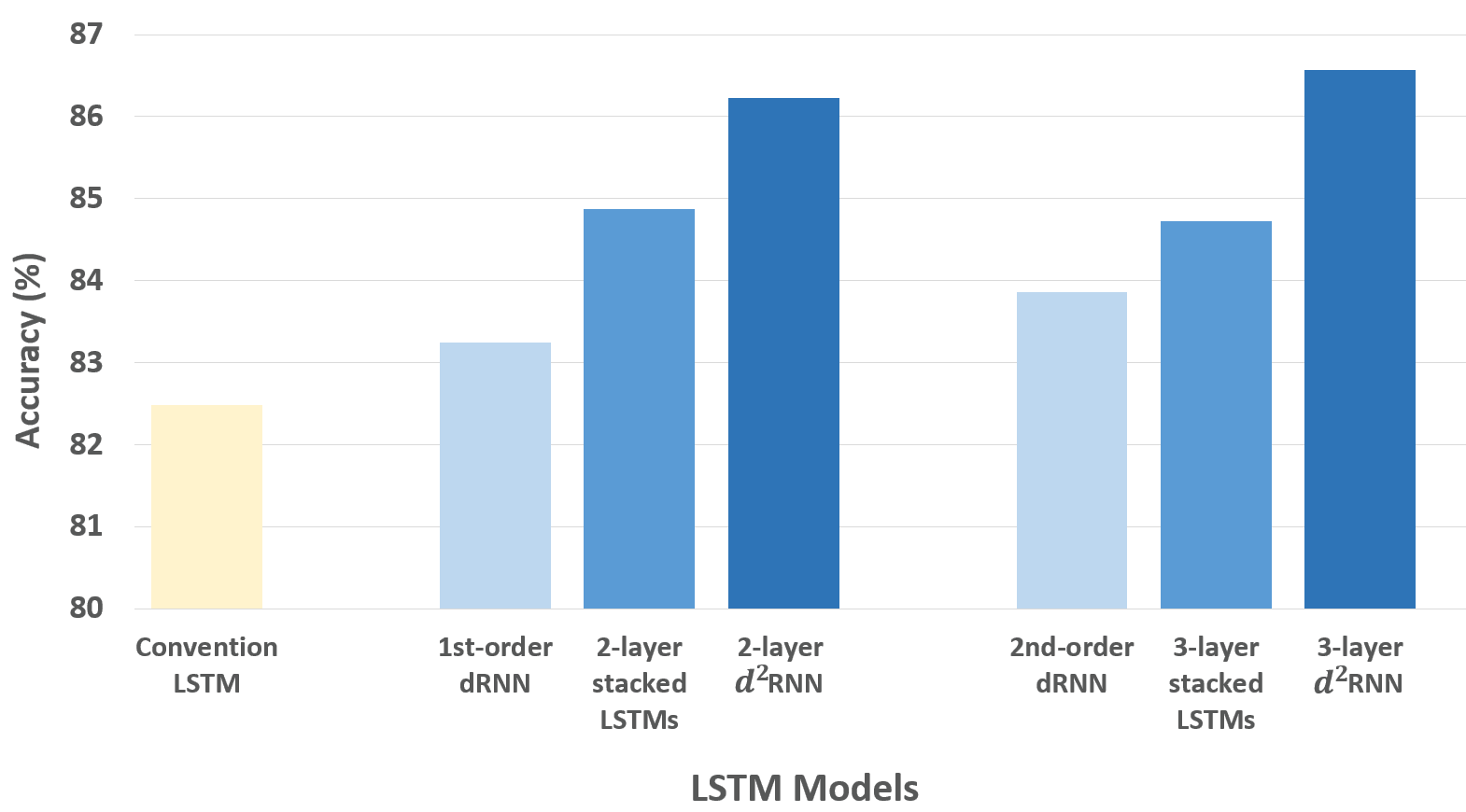}
	\caption{Comparison of the proposed $d^2$RNN model with other LSTM models on the Violent-Flows dataset.}
	\label{fig:vf_lstm}
\end{figure}
Figure \ref{fig:vf_lstm} compares the results of LSTM models on the Violent-Flows dataset. For the dRNN models, we report the accuracy up to the 2nd-order of DoS. For stacked LSTMs and $d^2$RNN, we report the accuracy up to 3 layers. 

From the results, dRNN, stacked LSTMs, and the proposed $d^2$RNN outperform the conventional LSTM algorithm with the same feature. $d^2$RNN outperforms both dRNN and stacked LSTMs, which demonstrates the effectiveness of learning intrinsic dynamical patterns present in video sequences with both deeper layers of LSTMs and higher-orders of DoS. While deeper layers enable greater model complexity, higher-orders of DoS strengthen the model's ability to detect salient spatio-temporal structures. In the meantime, the stacked deep layers in space naturally reveal the salient dynamics over time, and decrease the chance of misaligned DoS in different orders.
dRNN model suffers from the distortion of salient motion patterns due to the combination of different orders DoS. On the other hand, stacked LSTMs ignore the salient spatio-temporal structures, by simply stacking homogeneous layers onto the model.

It is worth pointing out that for stacked LSTMs and $d^2$RNN, the 3-layer architectures do not improve the performance much compared to their corresponding 2-layer models on the Violent-Flows dataset. For stacked LSTMs, the performance is even slightly decreasing. This is probably because with deep layers in the architecture, the model is getting high in complexity and has larger chance of overfitting.

\begin{figure}
	\centering
		\includegraphics[width=1.0\linewidth]{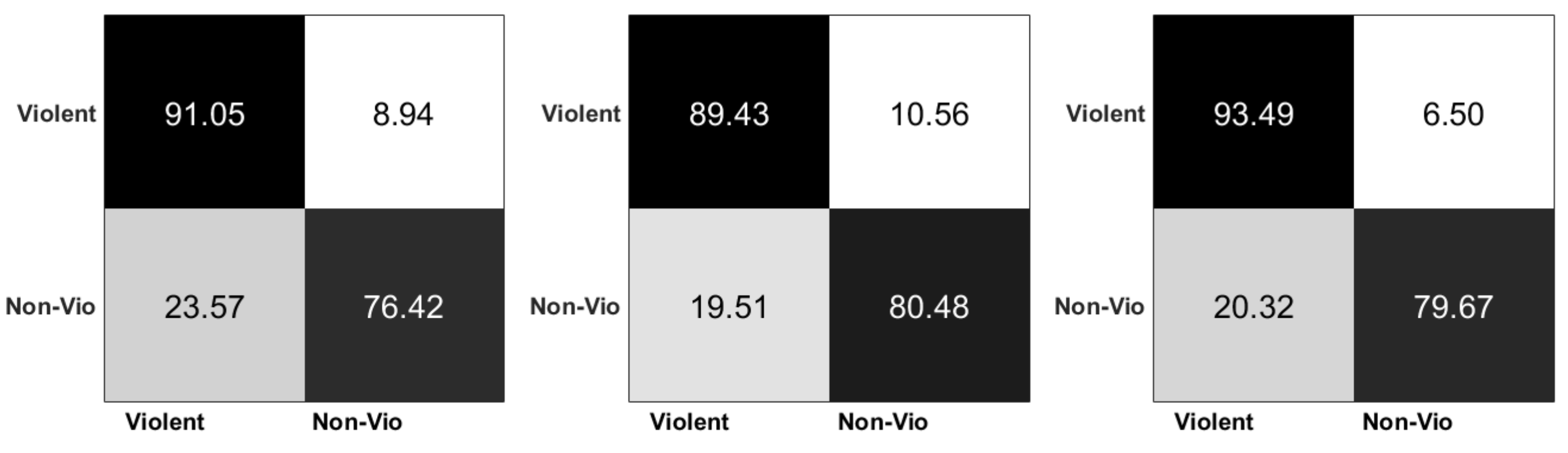}
	\caption{Confusion Matrices obtained by 2nd-order dRNN (left), 3-layer stacked LSTMs (middle), and 3-layer $d^2$RNN (right) on the Violent-Flows dataset }
	\label{fig:vf_cf}
\end{figure}

From Figure \ref{fig:vf_cf}, we can see that $d^2$RNNN can effectively detect the violent scenes, which demonstrates the superiority of using individual orders of DoS. Stacked LSTMs, without using DoS, perform less designedly in recognizing violent activities. This is probably due to the lack of strength in detecting motion saliency.

In Table \ref{tab:VF}, we compare 3-layer $d^2$RNN with the non-LSTM state-of-the-art algorithms on the Violent-Flows dataset. $d^2$RNN model outperforms other state-of-the-art methods, which again demonstrates our model's effectiveness in learning more complex dynamical patterns via deep stacked layers and detecting spatio-temporal saliency via a hierarchy of DoS.

\begin{table}
	\centering
		\begin{tabular}{ c | c  }
		\hline
			\textbf{Methods} & \textbf{Accuracy (\%)} \\ 
        \hline 
		    Violent Flows \cite{hassner2012violent} & 81.30 \\
			Common Measure \cite{mousavi2015crowd} & 81.50 \\
			Hist of Tracklet \cite{mousavi2015analyzing} & 82.30 \\
			Substantial Derivative \cite{mohammadi2015violence} & 85.43 \\
            Holistic Features \cite{marsden2016holistic} & 85.53 \\
            3-layer $d^2$RNN & \textbf{86.58} \\
		\hline	
		\end{tabular}
	\caption{Performance comparison on the Violent-Flows dataset.}
	\label{tab:VF}
\end{table}

\section{Conclusion}
\label{sec:conclusion}

In this paper, we present a novel LSTM model deep differential Recurrent Neural Network ($d^2$RNN), which integrates stacked LSTMs and Derivative of State (DoS). Instead of simply stacking homogeneous LSTM layers, $d^2$RNN stacks multiple levels of LSTM cells with individual and increasing orders of DoS. Our model inherits the strength of stacked LSTMs to model more complex dynamical patterns than the conventional LSTM. In addition, it further enables the ability to detect salient spatio-temporal structures via the hierarchy of DoS. On the other hand, $d^2$RNN differs substantially from dRNN. Instead of using the combination of different orders of DoS, which has been shown suboptimal, our model modulates the LSTM gates with individual orders DoS and mitigate the problem of information distortion. We demonstrate our model's superiority on human activity datasets by showing that $d^2$RNN outperforms LSTM, dRNN, and stacked LSTMs. Even in comparison with the other state-of-the-art methods based on strong assumptions about the motion structure of activities being studied, the general-purpose $d^2$RNN model still demonstrates competitive performance. In the future work, we will explore the potential of $d^2$RNN in broader applications, such as speech recognition, music synthesis, online handwriting recognition, video captioning, and gesture recognition.

\bibliographystyle{ACM-Reference-Format}
\bibliography{sample-bibliography}

\end{document}